\pdfoutput=1
\documentclass[11pt]{article}

\usepackage[final]{acl}

\usepackage{times}
\usepackage{latexsym}

\usepackage[T1]{fontenc}
\usepackage[utf8]{inputenc}

\usepackage{microtype}

\usepackage{graphicx}

\usepackage{natbib}

\usepackage{lipsum}

\usepackage{algorithm}
\usepackage[noend]{algpseudocode}
\usepackage{amsmath}
\usepackage{amssymb}  % Contains \mathit and other math symbols

\usepackage{booktabs}
\usepackage{adjustbox}
\newcommand{\ra}[1]{\renewcommand{\arraystretch}{#1}}
\algnewcommand\algorithmicinput{\textbf{Input:}}
\algnewcommand\algorithmicoutput{\textbf{Output:}}
\algnewcommand\Input[1]{\State \algorithmicinput\ #1}
\algnewcommand\Output[1]{\State \algorithmicoutput\ #1}
\usepackage{xcolor}

\title{
Train It and Forget It: Merge Lists are Unnecessary for BPE Inference in Language Models
}
\author{Tomohiro Sawada \thanks{Corresponding author: tsawada@gatech.edu.}, Kartik Goyal \\
Georgia Institute of Technology
}

\begin{document}
\maketitle

\begin{abstract}
Standard Byte-Pair Encoding (BPE) tokenization compresses text by pairing a learned token vocabulary with a detailed merge list. Recent work has shown that this merge list exposes a potential attack surface for extracting information about language model's training data.
In this paper, we explore the downstream impact of BPE inference algorithms that do not rely on this merge list at all, and hence differ from the encoding process during BPE training.
To address this question, we investigate two broad classes of BPE inference schemes that differ from BPE application during training: a) targeted deviation from merge-lists including random merge orders, and various corruptions of merge list involving deletion/truncation, and b) non-targeted BPE inference algorithms that do not depend on the merge list but focus on compressing the text either greedily or exactly. Extensive experiments across diverse language modeling tasks like accuracy-based QA benchmarks, machine translation, and open-ended generation reveal that while targeted deviation from the merge lists exhibits significant degradation in language model performance, the non-targeted merge-list-free inference algorithms result in minimal impact on downstream performance that is often much smaller than expected.
These findings pave way for simpler and potentially more privacy-preserving tokenization schemes that do not catastrophically compromise model performance.
\end{abstract}

\section{Introduction}
Byte-pair encoding \cite{gage_new_1994,sennrich-etal-2016-neural,kudo_sentencepiece_2018,radford_language_nodate}
is the standard algorithm used to tokenize input texts for large language models (LLMs). In practice, most BPE-based tokenizer implementations used for frontier language models\footnote{
Most notably, the Hugging Face tokenizer codebase: 
    \href{https://github.com/huggingface/tokenizers}{https://github.com/huggingface/tokenizers}
} rely on a learned merge list to iteratively combine subword units into tokens during inference time. This BPE inference procedure is appealing because it mimics the merge application procedure during BPE training. However, dependence on the learned merge list exposes a vulnerability that might facilitate exploits to affect the model's downstream performance. Also, as shown in recent work \cite{hayase_data_nodate}, these merge lists expose an attack surface where adversaries can steal information about the tokenizer's training data that is likely correlated with the LLM training data. Moreover, other works \cite{geiping2024coercingllmsrevealalmost} have shown that discrepancies between the tokenizer and LLM's training data can lead to "glitch tokens" which lead to generation failures thus, information about the tokenizer's training data can be used to finding and exploiting these glitches \cite{land2024fishingmagikarpautomaticallydetecting}. It is therefore undesirable to rely on the BPE merge list during the deployment of the associated language model.
% Add this to your preamble:
%\setlength{\tabcolsep}{4pt}
\begin{figure*}[ht]
\centering
{\footnotesize
    \begin{adjustbox}{max width=\textwidth}
    \ra{1.3}
    \includegraphics[width=\textwidth]{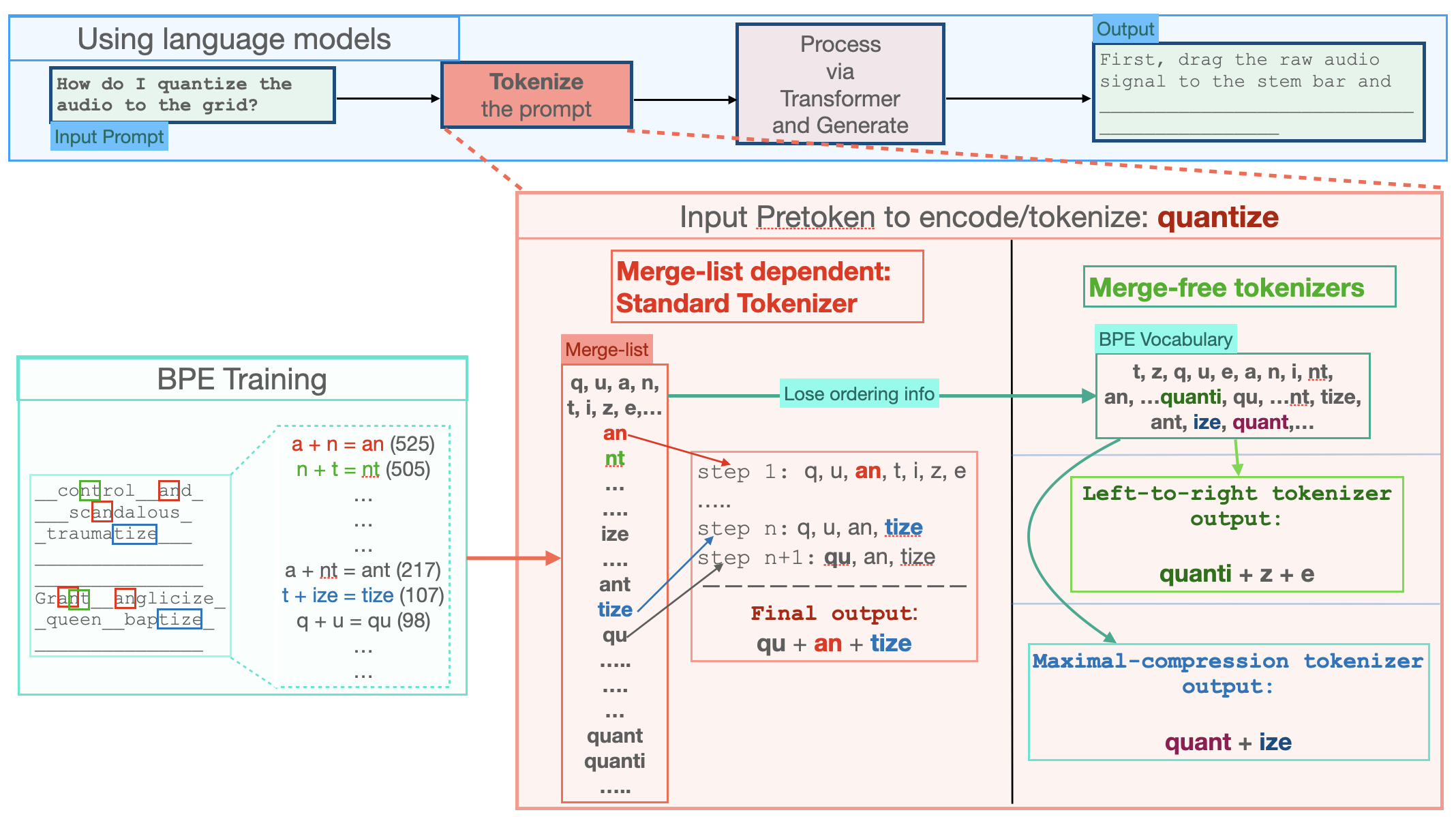}
\end{adjustbox}
}
\caption{
    Illustration comparing merge-list-based and merge-list-free BPE algorithms elaborated in the pink expanded box. The pretoken ``\emph{quantize}'' is tokenized by three different algorithms: a merge-list-based \texttt{standard} tokenizer (left) and two merge-list-free algorithms \texttt{left-to-right} (right-top) and \texttt{maximal-compression} (right-bottom). The ordered merge list is obtained from bigram statistics during BPE training. In contrast, merge-list-free algorithms only depend on the unordered BPE vocabulary, which contains less information about the training corpus.}
   % We show that merge-list free encoding algorithms perform just as well as the standard encoding algorithms, 
   % \textit{even if the language model was trained using the standard encoding algorithm} (training/inference mismatch).
%}
\label{fig:intro-diagram}
\end{figure*}

%and potentially expose details about the tokenizer’s, and potentially the LLM’s training data. Recent works  have shown that this merge list exposes an  
%Moreover, discrepancies between the merge list and the model’s training distribution can give rise to “glitch tokens” that degrade generation quality. Crucially, the original BPE formulation does not require maintaining or using merge lists at inference time. 
Hence in this paper, we investigate the effectiveness of using alternative BPE inference algorithms that do not depend on the learned merge lists post hoc for large language models trained with merge-list dependent BPE tokenization. BPE vocabulary typically admits multiple possible segmentations of the input pretokens that can be obtained from a myriad of BPE inference schemes. However, as we show in our experiments, these schemes are not all equal and the standard merge-list dependent scheme is ideal because of its alignment with the BPE training procedure.\footnote{Technically, the inference scheme used for tokenization of data during training of language models is the most ideal scheme. But in our experiments and general practice, the language models use the merge-list dependent BPE inference scheme.} Specifically, we focus on two such algorithms that aim to optimally compress the input text: a) left-to-right encoding that greedily maximizes compression; and b) an exact maximal compression encoding algorithm to compress the input pretokens given the BPE vocabulary. We contrast the impact of these algorithms to a class of inference algorithms that arise by targeted manipulation of the vulnerable merge list which includes truncation/deletion of merges, random shuffling of ordered merges, and backing-off to single characters. On three diverse language modeling tasks -- a) multiple-choice QA, b) conditional generation (machine translation), and c) open-ended generation -- we observe that the targeted inference algorithms significantly degrade the downstream LLM performance, but the non-targeted algorithms focusing on compression do not negatively impact LLM performance, and even improve it in some cases. Finally, we conduct further quantitative and qualitative analysis to study this surprising pattern of results in greater detail.

Our contributions are: i) empirically support compression-focused inference algorithms for tokenization which ameliorate the security vulnerability arising from the dependence on merge-lists; ii) investigate the downstream effect of numerous BPE inference algorithms, including ones that exploit the merge-list vulnerability, that deviate from training on diverse language modeling tasks; and iii) shed light on the extent to which the non-deterministic encoding property of BPE documented in prior work is impactful in practice.
%Byte-pair encoding (\cite{gage_new_1994,sennrich_neural_2016,kudo_sentencepiece_2018,radford_language_nodate})
%is the standard algorithm used to tokenize input texts for large language models (LLMs). 
%In many implementations used in practice
%\footnote{
%    Most notably, the Huggingface tokenizer codebase: 
%    \href{https://github.com/huggingface/tokenizers}{https://github.com/huggingface/tokenizers}
%}
%, the tokenizer encodes a pretoken by first checking for exact matches in the vocabulary, and 
%subsequently looking for elements in the merge list (learnt during training), and 
%applying the merge operations until the input is fully encoded.  

%In this work, we propose \textit{merge-list-free} BPE encoding algorithms where 
%given a BPE vocabulary, we encode a pretoken without relying on the merge list at all. 
%Furthermore, we claim that we can apply these changes post-hoc to an existing LLM: 
%given a model that was trained using a merge-based BPE encoding, 
%we can use merge-free encoding at inference time \textit{without} negatively impacting generation quality.
%We also show through our ablations that generation quality is 
%much more sensitive to the merge lists compared to the vocabulary.

\section{Training and Inference for BPE}
% We provide prerequisites for understanding our main contributions. 
% \tom{Emphasize the importance of \textit{compression} to downstream performance. This is the key motivation for left-to-right and maximal compression tokenizers.}
\label{sec:background}

Given a fixed BPE vocabulary, there are many possible encoding algorithms one can use to encode a pretoken.
\footnote{
    Given a BPE vocabulary, there are multiple ways to encode a given pretoken that are each deterministic functions of the input. 
    Following the terminology in \cite{gastaldi2025the}, we call such tokenizers \textit{non-deterministic}.  
}
Typically, BPE training produces merge-lists and inference also uses these merge lists in the same way as training to avoid mismatch and reduce ambiguity in segmentation. A merge-list-based BPE encoder is deterministic once the merge list is fixed; a vocabulary alone only specifies a family of possible deterministic encoders. Our focus in this paper is to explore \emph{mismatched inference algorithms for BPE on a model pretrained with a merge-list-based inference scheme}. 
%Since large language models are trained on sequences of tokens, 
%the tokenization process is a crucial component of the inference pipeline. 
In this section, we review how standard BPE training and encoding process and describe the two alternate merge-free BPE inference algorithms explored in this paper.

\subsection{Training and Merge-list}
%Byte-pair encoding (BPE)
%is a popular compression algorithm 
%commonly used in natural language processing. 
BPE is a greedy compression algorithm that is trained on a corpus by repeatedly merging the most frequent pair of tokens
in the training corpora, and recording the new merged token at each step into the BPE tokenizer vocabulary. In practice, each pretoken (space separated word) is processed individually across the corpus. This results in the vocabulary of the tokenizer. 

A lesser known fact is that many standard BPE implementations 
also record the merge list, which is the ordered list of merges that were performed sequentially during the training process (see Figure~1). 
This list has strictly more information than the vocabulary alone because it contains the 
"training dynamics" of the tokenizer, namely 
a.) the splittings of the tokens (and hence the "dependencies" between them), and 
b.) the order of the merges. 
Recent work \cite{hayase_data_nodate} has shown that this information can be used to 
extract information about the tokenizer's training data, which is often correlated with the 
pretraining data of the language model. 
Thus, the tokenizer merge lists are potential attack surfaces 
which adversaries can exploit to extract information about the language model. 
In contrast, the BPE vocabulary does not include any information about the order of the merges, 
and is more difficult to use for attacks.  

\subsection{Merge-based BPE Encoding Algorithms}
Standard implementations of BPE encodings 
use the merge list to encode pretokens returned by some pretokenization pipeline (which often returns a list of pretokens). To the best of our knowledge, there is no single consensus “standard BPE inference algorithm,” even though popular libraries follow merge-list-based inference variants, e.g., Hugging Face Tokenizers\footnote{\url{https://github.com/huggingface/tokenizers/blob/ee2c5708bdce9d6610fa74faeb22cf6297c6390a/tokenizers/src/models/bpe/model.rs\#L382C5-L468C1}}, tiktoken\footnote{\url{https://github.com/openai/tiktoken/blob/4560a8896f5fb1d35c6f8fd6eee0399f9a1a27ca/src/lib.rs\#L17-L83}}, SentencePiece (BPE mode)\footnote{\url{https://github.com/google/sentencepiece/blob/273449044caa593c2fd7eb7550cb3ab2cff93f1a/src/bpe_model.cc\#L38-L202}}, and fastBPE\footnote{\url{https://github.com/glample/fastBPE/blob/036711f8fdc3265d64e8e123a0761be12c5a8e74/fastBPE/fastBPE.hpp\#L581-L630}}.
The tokenizer first attempts to match the pretoken with an element in the vocabulary. 
If there are no exact matches, the tokenizer then takes a list of merges from the merge list appearing 
in the pretoken, and subsequently applies the merges to the pretoken as illustrated in Figure~1. The primary motivation behind this scheme is to emulate the same compression process at inference time as in 
the training process so that 
the token distribution seen by the models at inference time is similar to the training distribution. 
In this paper, we call the algorithm described above the \textit{merge-based} BPE encoding algorithm 
since it relies on the merge list at test time. 

An important aspect of merge lists is their natural \textit{hierarchical structure}. 
For example, if the bigram "an" is learnt at the first step of training, and the token "ant" is learnt at the seventh step by merging "an" and "t", then the token "ant" can only be used after applying the merge "a n", and so "ant" \textit{is a child of} "an". 
This is a key property of merge lists. We revisit this in our merge-list perturbation based experiments -- when we delete a symbol from the merge list, we must also delete all its children since they are no longer reachable during the standard BPE encoding process. 
%Why are BPE encoders implemented in this way?  
%The primary motivation is to emulate the same compression process at test time as in 
%the training process so that 
%the token distribution seen by the models at inference time is similar to the training distribution. 
As noted above, the merge lists provide a security risk which can have severe consequences to model 
providers. 
Our work shows that it is possible to encode text by patching this vulnerability 
\textit{while maintaining downstream performance}.
Moreover, our method does \textit{not} require retraining the language model on the new tokenizer, and 
can be applied post-hoc to any existing language model.

\subsection{Non-targeted merge-list-free BPE inference algorithms}

Given a BPE vocabulary, we can encode a pretoken without relying on the merge list. We explore two merge-list-free algorithms that focus on compression and integrate cleanly into the LM inference pipeline. These likely behave well because BPE training can be interpreted~\cite{zouhar-etal-2023-formal} as implicitly prioritizing compression in a greedy manner. We call them \emph{non-targeted} because they do not manipulate the learned merge list.
In this paper, we call a merge-list-free encoding algorithm \textit{performant} if it achieves 
comparable or better downstream performance as the standard merge-based encoding algorithm.

\subsubsection{Left-to-right greedy encoding}

The left-to-right encoding algorithm is a simple and efficient procedure for encoding a pretoken.
Given a pretoken, we look for the longest prefix of the pretoken that is in the vocabulary, 
and we output that prefix as a token. 
We then repeat this process for the remaining suffix of the pretoken. For example, given the pretoken "quantize" and the vocabulary provided in \ref{fig:intro-diagram}, 
the left-to-right encoding algorithm 
chooses the token "quanti" (as opposed to "quant" or "qu") since it is the longest prefix in the vocabulary. 
The suffix "ze" is then encoded as "z" and "e" since the string "ze" is not in the vocabulary.

This is a natural candidate for a performant merge-list-free encoder. Since BPE learns tokens greedily during training, it is plausible that left-to-right encoding achieves a similar level of compression.

\subsubsection{Maximal compression encoding}

Prior work \cite{goldman_unpacking_2024} has shown that compression during 
LLM \textit{pretraining} correlates strongly with downstream performance.
It is therefore natural to ask whether better compression at inference time leads to better downstream performance.
To address this question, we consider the \textit{maximal compression encoding algorithm}. 
Given a pretoken, we look for the combination of tokens in the vocabulary which gives the highest compression of the pretoken. 

For example, if we have the pretoken "quantize" and the vocabulary provided in \ref{fig:intro-diagram}, 
the string "quantize" is not an element in the vocabulary, so the shortest encoding must contain at least two tokens. 
From manual inspection, we see that "quant" and "ize" are both in the vocabulary, so the maximal compression encoding algorithm 
chooses this split.

A naive implementation has exponential time complexity in the pretoken length, but dynamic programming reduces this to quadratic time \footnote{See Appendix, Algorithm~\ref{algo:maxcompbpe}.}. In practice, pretokens are short after pretokenization.
%See \cite{uzan_greed_2024} for a more detailed discussion on the comparison between the left-to-right greedy encoding algorithm, maximal compression encoding algorithm, with other merge-list free encoding algorithms.

\subsection{Other merge-list-free encoding algorithms}

Although there are many other merge-free inference algorithms, many of them do not compress the prompt as well as the ones discussed above. 
The most trivial one is the character-based encoding algorithm: 
this breaks the pretoken into characters and outputs them as tokens. 
This encoding method has the \textit{worst} compression for a given piece of prompt, 
and is thus the opposite of the maximal compression encoding algorithm. 

As described in the subsequent sections, we observe that the compression-oriented inference algorithms, especially the left-to-right greedy encoding algorithm, 
have comparable downstream performance to the standard merge-based encoding algorithm, while the character-based encoding algorithm, although also merge-free, performs significantly worse.

\section{Impact of Training-Inference Mismatch on LM Performance}\label{sec:mismatch}

In this section, we describe our empirical findings on the impact of different tokenization schemes on downstream LM performance. We not only compare the merge-free non-targeted compression-based inference algorithms to the standard tokenization algorithm described above, but we also investigate other tokenization schemes that explicitly seek to exploit and manipulate the vulnerabilities offered by a publicly available merge-list. We perform extensive investigation on three diverse LM-based tasks as described below. The central question we aim to explore is the nature of the impact of the mismatch between training and inference time tokenization procedures. 
\subsection{Experimental Setup}

We evaluate an LLM on three diverse kinds of tasks: multiple-choice QA tasks that require very short form generation after encoding the question prompt, a longer conditional generation task of machine translation that involves processing a prompt with the source text and generating target text, and a fully open-ended generation task that focuses on completion based on context to be encoded. We process the prompts with the different encoding schemes, but generate with the full vocabulary. It must be noted that the choice of tokenization inference does not affect generation with a BPE-based tokenizer.

We choose to focus on the \texttt{Qwen-2-7B-Instruct} model \cite{yang2024qwen2technicalreport} for our experiments. 
The choice of model is motivated by the need for a model with a sizable vocabulary size to experiment with different ranges of corruptions) and 
a tokenizer which was trained using the HF tokenizer (as opposed to tiktoken).
\footnote{
    These desiderata eliminates other popular models such as OLMo-7B and Llama-3.
    The former uses the GPT-NeoX tokenizer, which has 50k tokens in its vocabulary,
    and the latter uses the tiktoken tokenizer. 

    The tiktoken tokenizer is a proprietary tokenizer developed by OpenAI for their models. 
    Their merge lists do not strictly adhere to the requirements we described in the previous section. 
}
The Qwen-2 tokenizer has 151645 tokens in its vocabulary, of which 255 are single character tokens. 
Unless noted otherwise, the Qwen-2 tokenizer will be referred to as the "standard" tokenizer 
(as opposed to the "custom" tokenizers obtained by either using a different encoding algorithm or by corrupting the merge list).
\subsubsection{MCQA tasks}
For the accuracy-based tasks, we evaluate the model on two popular Q\&A benchmarks: 
MMLU \cite{hendrycks_measuring_2021} and ARC-Easy/Challenge \cite{clark_think_2018}. 
\subsubsection{Conditional Generation: Machine Translation}
We consider the effect of different tokenizations on 
the semantic correctness of the generated text by testing it on the task of machine translation. 
We evaluate the performance of the model on WMT16 Czech→English and WMT15 German→English test sets, and report COMET \cite{rei_comet_2020} as our primary metric in the main paper. We relegate BLEU \cite{papineni_bleu_2002} and METEOR \cite{banerjee_meteor_2005}, along with xCOMET and CometKiwi, to the Appendix for completeness. BLEU is computed on detokenized outputs.
\subsubsection{Open-ended Generation}
We evaluated the open-ended generation capabilities of the model 
 by prompting it with abstracts from scientific papers. 
 The prompts were generated by extracting the first five sentences
from the full text in the "Semantic Scholar Open Research Corpus" \cite{lo_s2orc_2020}, 
a text corpus consisting of research papers extracted from Semantic Scholar. 
 To ensure the quality and diversity of the prompts, we took 5,899 examples from the corpus 
across multiple academic fields.  
The text corpora was chosen due to the high concentration of domain-specific pretokens 
which are likely to be sensitive to tokenization. 

 We then measured how much the generated text deviated from the 
 original human-written distribution by measuring the MAUVE score \cite{pillutla_mauve_2023} between the 
two sets of texts.

%To evaluate the performance of the tokenizers, we use  
\subsection{Targeted Tokenization}
As mentioned above, we measure the impact on the downstream LM performance when the inference algorithms target to manipulate merge-list obtained via tokenizer training. 
%By doing so, we can measure how much the corruptions 
%affect the semantic correctness of the generated text.
We deliberately corrupt the merge list of the tokenizer and measure the performance degradation. 
To understand the sensitivity of LLM inference on the merge list, 
we ran inference using tokenizations generated from a corrupted merge list. 
The merge list gives a fine-grained interface for controlling the encoding of the model 
(as opposed to the choice of encoding algorithms which are \textit{qualitatively} different from one another).
These experiments can also help us understand to what extent the manipulation of the merge list (by for example, a malicious insider)
can be used to sabotage the generation capability of the model. We corrupt the tokenizer in the following ways:

\noindent \textbf{Truncation}: Since the merge lists are generated in the order in which the merges are learned, 
we consider the effect of removing the less common merges (learned last during training) by 
deleting the last \textit{N} merges from the merge list.

\noindent \textbf{Deletion}: We also consider the effect of \textit{random deletion} of merges since 
the merges important for downstream performance may not be 
concentrated in a particular region within the merge list.
For random deletions, we first choose an initial set of deletions (the "initial set") 
and delete all merges which depend on these seeds (the "number of deletions"). 
%We report both of these numbers because the latter is the "effective" number of deletions 
%that resulted from deleting the initial set. 
To generate our random deletion tokenizers, we've fixed a random seed, chose an increasing number of initial deletions,  
and measured the performance of the model for each of these settings. 
(This is why the number of deletions is not a clean number for all of our random deletion experiments.)

\noindent \textbf{Merge Shuffle}: We also consider a merge-based tokenization where at runtime, we 
randomly shuffle the merge list being applied to the pretoken. 
For example, the standard encoding algorithm \ref{fig:intro-diagram} may tokenize the pretoken "quantize" 
by successively applying the merges "a n", "z e", "i ze", "t ize", and "q u", in this order, 
resulting in the tokenization "qu an tize". 
The random shuffle encoding algorithm may instead apply the merges 
"u a", "n t", "q ua", "nt i", and "ze" (assuming all of these appear in the merge list somewhere), 
resulting in the tokenization "qua nti ze". 
Throughout our experiments, we have a fixed random seed which determines how the merge list is shuffled. 

The random shuffle encoding results in a drastically different 
token distribution at inference time compared to the standard encoding algorithm. 
This provides a natural baseline where we expect the generation capability of the model 
to be significantly degraded.

\noindent \textbf{Character Level}: As described above, we also consider the baseline of splitting pretokens into individual characters.

% Version: 0519-0350

\setlength{\tabcolsep}{4pt}

\begin{table}[ht]
  \centering
  \begin{adjustbox}{width=0.47\textwidth}
    \ra{1.3}
    \begin{tabular}{@{}lrrrrr@{}}
      \toprule
       &  \multicolumn{2}{c}{Accuracy-based Tasks} 
       &  \multicolumn{2}{c}{Machine Translation} 
       &  \multicolumn{1}{c}{OEG.} \\
      \cmidrule(lr){2-3} \cmidrule(lr){4-5} \cmidrule(lr){6-6}
       Tokenizer        & ARC   & MMLU  & De$\rightarrow$En  & Cz$\rightarrow$En & MAUVE \\
      \midrule
       Standard             & 0.869 & 0.656 & 0.502 & 0.685  & 0.904 \\
       Merge shuffle        & 0.853 & 0.617 & 0.478 & 0.633  & 0.245 \\
       Character-level      & 0.860 & 0.624 & 0.479 & 0.520  & 0.399 \\
       Random deletion      & 0.860 & 0.628 & 0.483 & 0.531  & 0.170 \\
      \bottomrule
    \end{tabular}
  \end{adjustbox}
  \caption{
    Evaluation results for merge-list-based corruption tokenizers on the accuracy-based tasks (ARC and MMLU)
    and machine translation (COMET for WMT De$\rightarrow$En and Cz$\rightarrow$En). 
    The corrupted tokenizers do not suffer as much for accuracy-based tasks compared to 
    longer generation tasks. The random deletion tokenizer was obtained by 
    randomly deleting 149\,802 merges from the standard tokenizer's merge list. 
    “OEG.” stands for “Open-ended Generation.”}
  \label{tab:corrupted}
\end{table}

Observing the results in Table~1, we see that the corruption doesn't seem to affect the MCQA tasks much but it shows significant degradation in MT and open-ended generation under corruption.  
Although the prompts in accuracy-based benchmarks are long enough to have 
different tokenizations under our scheme, the generation length is not long enough 
to show substantial differences in performance.
The merge shuffle corruption 
consistently performs at least as bad as, if not worse than, the character-level corruption. 
This suggests that severe corruption to the merge lists 
can essentially do away any benefits of subword tokenization, 
and the model may as well use a character-level tokenization.  

% Add this to your preamble:
\setlength{\tabcolsep}{4pt}
\begin{figure*}[ht]
\centering
{\footnotesize
    \begin{adjustbox}{max width=\textwidth}
    \ra{1.3}
    \includegraphics[width=0.8\textwidth]{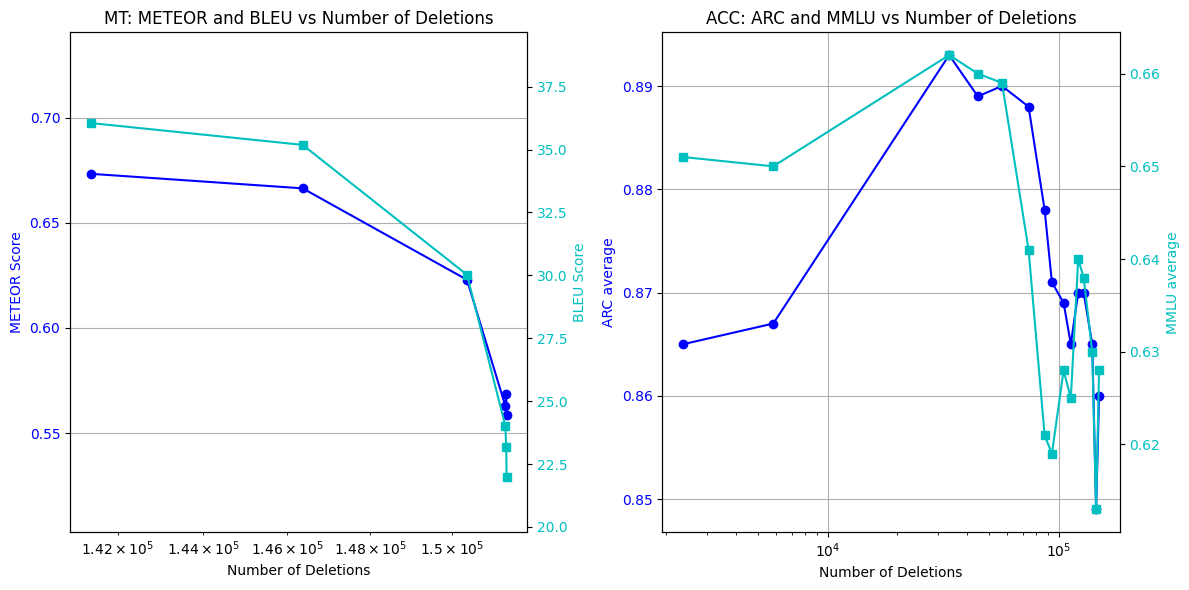}
    \end{adjustbox}
}
\caption{Performance of different random deletion tokenizers on 
the accuracy-based tasks (ARC and MMLU) and the machine translation task. 
For both datasets, the performance degrades after around 70k deletions. 
}
\label{fig:edit}
\end{figure*}
In Figure~2, we investigate the relationship between the effect on downstream performance and severity of corruption. We observe that both semantic and n-gram metrics are not too sensitive to mild corruption 
on a per-example level. As the corruption levels cross a threshold (merge shuffle, char-level, aggressive deletion), the drop in performance is noticeably significant. 
In fact, we've observed that performance is quite stable even for "medium-sized" deletions (107060 and 115604). 
This seems to suggest that the model's performance relies primarily on "highly-trained" tokens
which are only destroyed for very aggressive corruptions. It could also be the case that large portions of BPE vocabulary are never used for practical purposes indicating the existence of many undertrained tokens in the vocabulary.
It is also interesting to note that the decline for the random deletion tokenizer 
is more steady in the machine translation task
compared to the accuracy-based tasks. 
This robustness is likely due to the fact that the model is generating longer text 
in the machine translation task. 

Overall on manual inspection, the degradation of generated output exhibits unnatural syntactic choices (e.g., characters separated out by spaces) which causes drops in BLEU and MAUVE.

\subsection{Non-targeted Tokenization}
As described above, we compare compression-based merge-free algorithms against the standard algorithm. These algorithms either greedily or exactly maximize compression of the pretoken given the BPE vocabulary.
% Version: 0519-0350

\setlength{\tabcolsep}{4pt}

\begin{table}[h]
  \centering
  \begin{adjustbox}{width=0.47\textwidth}
    \ra{1.3}
    \begin{tabular}{@{}lrrrrr@{}}
      \toprule
       &  \multicolumn{2}{c}{Accuracy-based Tasks} 
       &  \multicolumn{2}{c}{Machine Translation} 
       &  \multicolumn{1}{c}{OEG.} \\
      \cmidrule(lr){2-3} \cmidrule(lr){4-5} \cmidrule(lr){6-6}
       Tokenizer          & ARC   & MMLU  & De$\rightarrow$En  & Cz$\rightarrow$En & MAUVE \\
      \midrule
       Standard             & 0.869 & 0.656 & 0.502 & 0.685 & 0.904 \\
       Maximal Compression  & 0.863 & 0.678 & 0.494 & 0.633 & 0.927 \\
       Left to right        & 0.903 & 0.705 & 0.495 & 0.633 & 0.985 \\
      \bottomrule
    \end{tabular}
  \end{adjustbox}
  \caption{
    Evaluation results for merge-list-free tokenizers on the accuracy-based tasks (ARC and MMLU), machine translation (COMET for WMT De$\rightarrow$En and Cz$\rightarrow$En),
     and the open-ended generation task. 
     The left-to-right tokenizer maintains performance or even outperforms the standard tokenizer on QA and OEG; small degradations are observed on MT (see text). The maximal compression also largely maintains the standard tokenizer's performance.
    “OEG.” stands for “Open-ended Generation.”}
  \label{tab:acc-mt}
\end{table}

In Table~\ref{tab:acc-mt}, we observe that left-to-right and maximal compression tokenization schemes maintain performance on accuracy-based QA and OEG, with left-to-right even improving ARC/MMLU and MAUVE. For machine translation, however, COMET shows modest but consistent drops relative to the standard tokenizer: for De$\rightarrow$En, 0.5017 (standard) vs. 0.4953 (left-to-right) and 0.4944 (maximal); for Cz$\rightarrow$En, 0.6853 (standard) vs. 0.6325/0.6328. These differences are potentially meaningful, particularly for Cz$\rightarrow$En. In Appendix Table~\ref{tab:tokenizer-comparison} and Table~\ref{tab:de-to-en-comet}, BLEU and METEOR broadly reflect the same directionality, though COMET appears more sensitive on Cz$\rightarrow$En. This aligns with COMET’s semantic focus, while BLEU/METEOR capture n-gram surface deviations. 
Overall, these results indicate that merge-list-free, compression-based encoders are robust across tasks but can induce small MT degradations, especially in morphologically richer settings.
% Add this to your preamble:
\setlength{\tabcolsep}{4pt}

\begin{table*}[ht]
  \centering
  {\footnotesize
    \begin{adjustbox}{max width=\textwidth}
      \ra{1.3}
      \begin{tabular}{@{}l rrrr@{}}
        \toprule
        Tokenizer       & Jaccard & Levenshtein & Edit & Perplexity \\
        \midrule  
        Standard        & 0.000  & 0.000      & 0.000 & 83.798 \\
        Left to right   & 0.226  & 29.645     & 0.165 &  95.891\\
        Maximal Comp.   & 0.196  & 24.740     & 0.139 & 155.751 \\
        Merge Shuffle   & 0.918  & 692.000     & 0.959 & 131.400 \\
        Character-level & 0.925  & 796.987     & 0.964 & 58.212 \\
        Random Deletion & 0.927  & 800.719     & 0.966 & 92.734 \\
        Truncation      & 0.889  & 455.775     & 0.884 & 97.202 \\
        \bottomrule
      \end{tabular}
    \end{adjustbox}
  }
  \caption{Perplexity scores and prompt metrics (Jaccard similarity, Levenshtein distance, edit distance) between different tokenization approaches and standard tokenization.}
  \label{tab:prompt-metrics}
\end{table*}

This is because when we measure how differently the prompts are encoded under various tokenization schemes compared to the standard tokenizer, we find that the merge-free tokenizers differ in the encoding of every single prompt in the open-ended generation task. Table~3 shows the average edit distances between the merge-based standard tokenizer encodings and encodings from the other tokenizers. We observe that the left-to-right and maximal compression encodings are less distant than other corruption-based tokenizers. Though, we also notice that they have higher perplexity on the prompts than the standard merge-based tokenizer. This indicates that the compression-based approaches use potentially unconventional and undertrained tokens, but these effects are overcome by the model's robustness to specific kinds of typos and over-segmentations associated with compression-based algorithms.

%For open-ended generation, merge-list free encoding methods yield 
%higher score than original tokenizer while for machine translation, there is a slight drop in performance. 

%\subsection{Open-ended generation.}

% \input{data/experimental_results/prompt-completion-combined}
% \input{data/experimental_results/mauve_plot-rand_del}

%While the open-ended generation task has a reference text like the machine translation task, 
%there are many valid completions for the same prompt which may vary substantially from the reference
%(hence "open-ended"). 
%We would like to investigate how the distribution of the completions shift under different encoding schemes. 

%\subsection{Merge shuffle is worse than character-level corruption.}

%Observe that for all of the above tasks, t 

%The remarkable performance of targeted tokenization agorithms (that is, left-to-right encoding and maximal compression) leads us to believe that 
%these tokenizers encode pretokens in a way similar to the original tokenizers. 
%However, this seems to not to be the case--
%encodings generated by either tokenizer has nontrivial 
%Levenstein distance with respect to the original tokenizer's encodings. 
%This suggests that these tokenizers are performant \textit{despite} having different encodings than the original tokenizer. 
%This leads us to ask: what changes in tokenization matter for downstream performance? 

\section{Characterizing Bad Tokenization}\label{sec:analysis}

We analyze why different tokenizations lead to different downstream performance, complementing the aggregate results in Section~\ref{sec:mismatch}. We present quantitative indicators of sensitivity/robustness and qualitative patterns in tokenization differences.

\subsection{Quantitative indicators of sensitivity and robustness}

\begin{table*}[t]
\centering
\ra{1.3}
\begin{tabular}{@{}lrrrr@{}}
\toprule
& \multicolumn{2}{c}{Cluster Size} & \multicolumn{2}{c}{Min Norm} \\
\cmidrule(lr){2-3} \cmidrule(lr){4-5}
& Robust  & Sensitive & Robust & Sensitive  \\
\midrule
Random deletion     & 420 & 157 & 2891.22 & 1609.89 \\
Character-level          & 430 & 115 & 2891.01 & 1606.73 \\
Merge shuffle       & 415 & 128 & 2837.74 & 1539.82 \\
Left to right       & 406 &  64 & 2748.35 & 1626.64 \\
Maximal compression & 397 &  60 & 2720.20 & 1597.86 \\
\bottomrule
\end{tabular}
\caption{
    Cluster size and minimum token embedding norm computed over each cluster. 
    We analyze the WMT15 de-en test set (2998 samples). 
    Notice that a) the robust cluster tends to be larger for non-targeted tokenizers compared to targeted tokenizers, and 
    b) the norm for sensitive cluster is consistently smaller than robust cluster. 
}
\label{tab:qualitative_analysis}
\end{table*}

Why do some tokenizations lead to better performance than others? One hypothesis is the presence of \textit{undertrained tokens}, i.e., vocabulary elements that occurred infrequently during pretraining. Past work \cite{land2024fishingmagikarpautomaticallydetecting} shows such tokens can cause undesirable behavior, e.g., failures to follow instructions. Following \cite{land2024fishingmagikarpautomaticallydetecting}, we use the minimum token-embedding norm as a proxy for undertrainedness.\footnote{If a token appears rarely during pretraining, its embedding sees fewer updates and tends to remain closer to the origin.}

Prior work \cite{bigelow2024forkingpathsneuraltext} also shows that changing a single token can drastically alter generations, suggesting even a small number of undertrained tokens in a prompt may lead to poor behavior. To study this, we examine “surprising” cases where large tokenization deviations yield small performance changes (robust) and where small tokenization deviations yield large performance changes (sensitive). For each generated sequence $L$, we compute

\begin{equation}
    \min_{t\in L} \, \lVert E(t) \rVert,
\end{equation}

where $\lVert\cdot\rVert$ is the norm and $E(t)$ is the token embedding. In Table~\ref{tab:qualitative_analysis}, sensitive clusters are consistently larger for targeted tokenizations than for merge-list-free ones, indicating targeted schemes more often induce small perturbations that cause large performance deviations. Sensitive clusters also exhibit smaller minimum norms than robust clusters, consistent with a higher incidence of undertrained tokens.

\subsection{Qualitative patterns in tokenization differences}

\begin{figure}[h]
\centering
\includegraphics[scale=0.38]{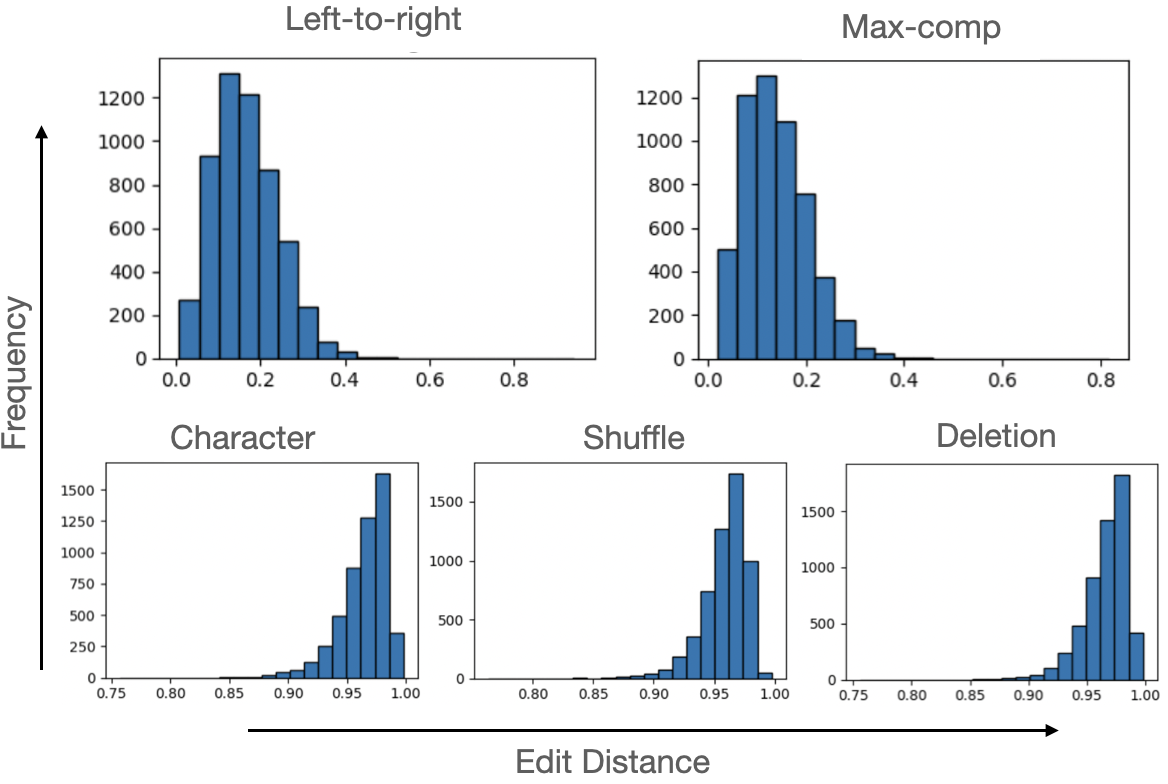}
\caption{Distribution of token-level edit distance between standard and alternative tokenizers for open-ended generation (S2ORC subset; N noted in text). Top: non-targeted merge-list-free schemes. Bottom: targeted schemes. Lower is closer to standard.}
\label{fig:edit-spread}
\end{figure}

We further examine token-level edit distance (Levenshtein distance between token-ID sequences) relative to the standard tokenizer for open-ended generation. Figure~\ref{fig:edit-spread} summarizes the distributions; merge-list-free (non-targeted) algorithms produce smaller encoding differences than targeted schemes.

Qualitatively, low-distance prompts cover diverse scientific domains and use simpler language; high-distance prompts are often biomedical with hyperspecific jargon and rare terms, consistent with more rarely trained tokens. This aligns with the quantitative evidence that sensitive cases are associated with smaller embedding norms.

\section{Related Work}
While we focus on BPE inference algorithms that ameliorate security vulnerabilities associated with merge-lists~\cite{hayase_data_nodate}, the non-deterministic property
\footnote{
    See \ref{sec:background} for the use of the term ``non-deterministic tokenization.''
}
of tokenization algorithms~\cite{kudo_sentencepiece_2018, sennrich-etal-2016-neural, mielke_between_2021} in general—identified in several prior works~\cite{cao-rimell-2021-evaluate, gastaldi2025the}—forms the crux of our motivation. The symbols in the vocabulary can give rise to multiple possible segmentations for a given word/pretoken. While much work has studied the effect of training different types of tokenizers/segmenters \emph{and} models based on those tokenizers~\cite{goldman-etal-2024-unpacking, saleva-lignos-2023-changes}, we instead focus on evaluating different BPE inference schemes on pretrained tokenizers and models with the standard BPE approach. While training models~\cite{provilkov-etal-2020-bpe} with different tokenization schemes in general does not affect downstream performance significantly, in our setting of training–inference mismatch we observe significant performance degradation with certain algorithms. Related to our work, \citet{uzan_greed_2024} also study different BPE inference algorithms, but they limit their analysis to intrinsic tokenization metrics like cognitive plausibility~\cite{beinborn-pinter-2023-analyzing} and morphology~\cite{bostrom-durrett-2020-byte} and do not investigate their downstream impact on model performance. Our finding that algorithms like left-to-right and maximal compression do not result in significant performance degradation despite encoding the prompts differently is also related to recent findings that LLMs have an implicit lexicon of pretokens~\cite{kaplan2025from} and are robust to typos~\cite{cao-etal-2023-unnatural}.
%\tom{Forking tokens paper.}
%\subsection{Tokenizer-based attacks on LLMs.}
%\tom{Talk about Magikarp and DMI.}

\section{Conclusion}
In light of security vulnerabilities associated with inference-time usage of the merge-list learned during BPE training, we explored alternative merge-free algorithms for BPE inference on pretrained models. We found that although arbitrary and targeted inference-time deviations from standard BPE hurt downstream LM performance significantly, surprisingly the non-targeted compression-based merge-free algorithms maintained or even improved it. This suggests potential overlap in the implicit objectives of BPE training and these merge-free algorithms paving way for more secure tokenization schemes for language models.
\section{Limitations}
The primary limitation of our work is that while we have articulated the need for merge-list-free BPE inference algorithms and have provided empirical evidence for two such inference algorithms focusing on compression across a diverse set of LM tasks, it is not clear that the algorithms investigated are the optimal algorithms for merge-free inference that preserves performance across \emph{all} domains and languages. Relatedly, we only have empirical support from our experiments and prior works for concluding that left-to-right and maximal compression algorithms preserve performance—possibly because the original BPE training procedure implicitly greedily optimizes~\cite{zouhar-etal-2023-formal} for compression and breaks ties in a left-to-right manner for most languages. We do not have theoretical support and guarantees for this conjecture, and our findings might not hold for small amounts of data in low-resource languages, especially with a non-monotonic or a non-left-to-right writing order. Finally, while our recommendation might eliminate data inference and other security vulnerabilities directly related to merge-lists, they still would not defend against other kinds of attacks based on tokenization such as those focusing on finding and exploiting \emph{glitch} tokens. 

\section{Ethical Considerations}
While we recommend defending against vulnerabilities associated with merge lists during deployment by not using them, this would also result in less transparency. It can be argued that publicly available merge-lists possible allow data-mixture inference and it might be desirable in certain cases because of transparency and auditability reasons. However, depending on the context, it can also be argued that LMs should be protected from the security vulnerabilities posed by publicly available merge-lists. We recognize that our recommendation applies for the latter contexts and doesn't apply in contexts that disproportionately prioritize transparency.
% Required by Call: https://aclrollingreview.org/cfp#long-papers

\bibliography{main}

\begin{thebibliography}{29}
\providecommand{\natexlab}[1]{#1}

\bibitem[{Banerjee and Lavie(2005)}]{banerjee_meteor_2005}
Satanjeev Banerjee and Alon Lavie. 2005.
\newblock \href {https://aclanthology.org/W05-0909/} {{METEOR}: {An} {Automatic} {Metric} for {MT} {Evaluation} with {Improved} {Correlation} with {Human} {Judgments}}.
\newblock In \emph{Proceedings of the {ACL} {Workshop} on {Intrinsic} and {Extrinsic} {Evaluation} {Measures} for {Machine} {Translation} and/or {Summarization}}, pages 65--72, Ann Arbor, Michigan. Association for Computational Linguistics.

\bibitem[{Beinborn and Pinter(2023)}]{beinborn-pinter-2023-analyzing}
Lisa Beinborn and Yuval Pinter. 2023.
\newblock \href {https://doi.org/10.18653/v1/2023.emnlp-main.272} {Analyzing cognitive plausibility of subword tokenization}.
\newblock In \emph{Proceedings of the 2023 Conference on Empirical Methods in Natural Language Processing}, pages 4478--4486, Singapore. Association for Computational Linguistics.

\bibitem[{Bigelow et~al.(2024)Bigelow, Holtzman, Tanaka, and Ullman}]{bigelow2024forkingpathsneuraltext}
Eric Bigelow, Ari Holtzman, Hidenori Tanaka, and Tomer Ullman. 2024.
\newblock \href {https://arxiv.org/abs/2412.07961} {Forking paths in neural text generation}.
\newblock \emph{Preprint}, arXiv:2412.07961.

\bibitem[{Bostrom and Durrett(2020)}]{bostrom-durrett-2020-byte}
Kaj Bostrom and Greg Durrett. 2020.
\newblock \href {https://doi.org/10.18653/v1/2020.findings-emnlp.414} {Byte pair encoding is suboptimal for language model pretraining}.
\newblock In \emph{Findings of the Association for Computational Linguistics: EMNLP 2020}, pages 4617--4624, Online. Association for Computational Linguistics.

\bibitem[{Cao and Rimell(2021)}]{cao-rimell-2021-evaluate}
Kris Cao and Laura Rimell. 2021.
\newblock \href {https://doi.org/10.18653/v1/2021.emnlp-main.161} {You should evaluate your language model on marginal likelihood over tokenisations}.
\newblock In \emph{Proceedings of the 2021 Conference on Empirical Methods in Natural Language Processing}, pages 2104--2114, Online and Punta Cana, Dominican Republic. Association for Computational Linguistics.

\bibitem[{Cao et~al.(2023)Cao, Kojima, Matsuo, and Iwasawa}]{cao-etal-2023-unnatural}
Qi~Cao, Takeshi Kojima, Yutaka Matsuo, and Yusuke Iwasawa. 2023.
\newblock \href {https://doi.org/10.18653/v1/2023.emnlp-main.550} {Unnatural error correction: {GPT}-4 can almost perfectly handle unnatural scrambled text}.
\newblock In \emph{Proceedings of the 2023 Conference on Empirical Methods in Natural Language Processing}, pages 8898--8913, Singapore. Association for Computational Linguistics.

\bibitem[{Clark et~al.(2018)Clark, Cowhey, Etzioni, Khot, Sabharwal, Schoenick, and Tafjord}]{clark_think_2018}
Peter Clark, Isaac Cowhey, Oren Etzioni, Tushar Khot, Ashish Sabharwal, Carissa Schoenick, and Oyvind Tafjord. 2018.
\newblock \href {https://doi.org/10.48550/arXiv.1803.05457} {Think you have {Solved} {Question} {Answering}? {Try} {ARC}, the {AI2} {Reasoning} {Challenge}}.
\newblock \emph{arXiv preprint}.
\newblock ArXiv:1803.05457 [cs] version: 1.

\bibitem[{Gage(1994)}]{gage_new_1994}
Philip Gage. 1994.
\newblock \href {https://www.semanticscholar.org/paper/A-new-algorithm-for-data-compression-Gage/1aa9c0045f1fe8c79cce03c7c14ef4b4643a21f8} {A new algorithm for data compression}.
\newblock \emph{The C Users Journal archive}.

\bibitem[{Gastaldi et~al.(2025)Gastaldi, Terilla, Malagutti, DuSell, Vieira, and Cotterell}]{gastaldi2025the}
Juan~Luis Gastaldi, John Terilla, Luca Malagutti, Brian DuSell, Tim Vieira, and Ryan Cotterell. 2025.
\newblock \href {https://openreview.net/forum?id=B5iOSxM2I0} {The foundations of tokenization: Statistical and computational concerns}.
\newblock In \emph{The Thirteenth International Conference on Learning Representations}.

\bibitem[{Geiping et~al.(2024)Geiping, Stein, Shu, Saifullah, Wen, and Goldstein}]{geiping2024coercingllmsrevealalmost}
Jonas Geiping, Alex Stein, Manli Shu, Khalid Saifullah, Yuxin Wen, and Tom Goldstein. 2024.
\newblock \href {https://arxiv.org/abs/2402.14020} {Coercing llms to do and reveal (almost) anything}.
\newblock \emph{Preprint}, arXiv:2402.14020.

\bibitem[{Goldman et~al.(2024{\natexlab{a}})Goldman, Caciularu, Eyal, Cao, Szpektor, and Tsarfaty}]{goldman_unpacking_2024}
Omer Goldman, Avi Caciularu, Matan Eyal, Kris Cao, Idan Szpektor, and Reut Tsarfaty. 2024{\natexlab{a}}.
\newblock \href {http://arxiv.org/abs/2403.06265} {Unpacking {Tokenization}: {Evaluating} {Text} {Compression} and its {Correlation} with {Model} {Performance}}.
\newblock \emph{arXiv preprint}.
\newblock ArXiv:2403.06265 [cs].

\bibitem[{Goldman et~al.(2024{\natexlab{b}})Goldman, Caciularu, Eyal, Cao, Szpektor, and Tsarfaty}]{goldman-etal-2024-unpacking}
Omer Goldman, Avi Caciularu, Matan Eyal, Kris Cao, Idan Szpektor, and Reut Tsarfaty. 2024{\natexlab{b}}.
\newblock \href {https://doi.org/10.18653/v1/2024.findings-acl.134} {Unpacking tokenization: Evaluating text compression and its correlation with model performance}.
\newblock In \emph{Findings of the Association for Computational Linguistics: ACL 2024}, pages 2274--2286, Bangkok, Thailand. Association for Computational Linguistics.

\bibitem[{Hayase et~al.(2024)Hayase, Liu, Choi, Oh, and Smith}]{hayase_data_nodate}
Jonathan Hayase, Alisa Liu, Yejin Choi, Sewoong Oh, and Noah~A Smith. 2024.
\newblock Data {Mixture} {Inference}: {What} do {BPE} {Tokenizers} {Reveal} about their {Training} {Data}?

\bibitem[{Hendrycks et~al.(2021)Hendrycks, Burns, Basart, Zou, Mazeika, Song, and Steinhardt}]{hendrycks_measuring_2021}
Dan Hendrycks, Collin Burns, Steven Basart, Andy Zou, Mantas Mazeika, Dawn Song, and Jacob Steinhardt. 2021.
\newblock \href {https://doi.org/10.48550/arXiv.2009.03300} {Measuring {Massive} {Multitask} {Language} {Understanding}}.
\newblock \emph{arXiv preprint}.
\newblock ArXiv:2009.03300 [cs].

\bibitem[{Kaplan et~al.(2025)Kaplan, Oren, Reif, and Schwartz}]{kaplan2025from}
Guy Kaplan, Matanel Oren, Yuval Reif, and Roy Schwartz. 2025.
\newblock \href {https://openreview.net/forum?id=328vch6tRs} {From tokens to words: On the inner lexicon of {LLM}s}.
\newblock In \emph{The Thirteenth International Conference on Learning Representations}.

\bibitem[{Kudo and Richardson(2018)}]{kudo_sentencepiece_2018}
Taku Kudo and John Richardson. 2018.
\newblock \href {https://doi.org/10.18653/v1/D18-2012} {{SentencePiece}: {A} simple and language independent subword tokenizer and detokenizer for {Neural} {Text} {Processing}}.
\newblock In \emph{Proceedings of the 2018 {Conference} on {Empirical} {Methods} in {Natural} {Language} {Processing}: {System} {Demonstrations}}, pages 66--71, Brussels, Belgium. Association for Computational Linguistics.

\bibitem[{Land and Bartolo(2024)}]{land2024fishingmagikarpautomaticallydetecting}
Sander Land and Max Bartolo. 2024.
\newblock \href {https://arxiv.org/abs/2405.05417} {Fishing for magikarp: Automatically detecting under-trained tokens in large language models}.
\newblock \emph{Preprint}, arXiv:2405.05417.

\bibitem[{Lo et~al.(2020)Lo, Wang, Neumann, Kinney, and Weld}]{lo_s2orc_2020}
Kyle Lo, Lucy~Lu Wang, Mark Neumann, Rodney Kinney, and Daniel Weld. 2020.
\newblock \href {https://doi.org/10.18653/v1/2020.acl-main.447} {{S2ORC}: {The} {Semantic} {Scholar} {Open} {Research} {Corpus}}.
\newblock In \emph{Proceedings of the 58th {Annual} {Meeting} of the {Association} for {Computational} {Linguistics}}, pages 4969--4983, Online. Association for Computational Linguistics.

\bibitem[{Mielke et~al.(2021)Mielke, Alyafeai, Salesky, Raffel, Dey, Gallé, Raja, Si, Lee, Sagot, and Tan}]{mielke_between_2021}
Sabrina~J. Mielke, Zaid Alyafeai, Elizabeth Salesky, Colin Raffel, Manan Dey, Matthias Gallé, Arun Raja, Chenglei Si, Wilson~Y. Lee, Benoît Sagot, and Samson Tan. 2021.
\newblock \href {https://doi.org/10.48550/arXiv.2112.10508} {Between words and characters: {A} {Brief} {History} of {Open}-{Vocabulary} {Modeling} and {Tokenization} in {NLP}}.
\newblock \emph{arXiv preprint}.
\newblock 2.

\bibitem[{Papineni et~al.(2002)Papineni, Roukos, Ward, and Zhu}]{papineni_bleu_2002}
Kishore Papineni, Salim Roukos, Todd Ward, and Wei-Jing Zhu. 2002.
\newblock \href {https://doi.org/10.3115/1073083.1073135} {Bleu: a {Method} for {Automatic} {Evaluation} of {Machine} {Translation}}.
\newblock In \emph{Proceedings of the 40th {Annual} {Meeting} of the {Association} for {Computational} {Linguistics}}, pages 311--318, Philadelphia, Pennsylvania, USA. Association for Computational Linguistics.

\bibitem[{Pillutla et~al.(2023)Pillutla, Liu, Thickstun, Welleck, Swayamdipta, Zellers, Oh, Choi, and Harchaoui}]{pillutla_mauve_2023}
Krishna Pillutla, Lang Liu, John Thickstun, Sean Welleck, Swabha Swayamdipta, Rowan Zellers, Sewoong Oh, Yejin Choi, and Zaid Harchaoui. 2023.
\newblock \href {https://doi.org/10.48550/arXiv.2212.14578} {{MAUVE} {Scores} for {Generative} {Models}: {Theory} and {Practice}}.
\newblock \emph{arXiv preprint}.
\newblock ArXiv:2212.14578 [cs].

\bibitem[{Provilkov et~al.(2020)Provilkov, Emelianenko, and Voita}]{provilkov-etal-2020-bpe}
Ivan Provilkov, Dmitrii Emelianenko, and Elena Voita. 2020.
\newblock \href {https://doi.org/10.18653/v1/2020.acl-main.170} {{BPE}-dropout: Simple and effective subword regularization}.
\newblock In \emph{Proceedings of the 58th Annual Meeting of the Association for Computational Linguistics}, pages 1882--1892, Online. Association for Computational Linguistics.

\bibitem[{Radford et~al.()Radford, Wu, Child, Luan, Amodei, and Sutskever}]{radford_language_nodate}
Alec Radford, Jeffrey Wu, Rewon Child, David Luan, Dario Amodei, and Ilya Sutskever.
\newblock Language {Models} are {Unsupervised} {Multitask} {Learners}.

\bibitem[{Rei et~al.(2020)Rei, Stewart, Farinha, and Lavie}]{rei_comet_2020}
Ricardo Rei, Craig Stewart, Ana~C Farinha, and Alon Lavie. 2020.
\newblock \href {https://doi.org/10.18653/v1/2020.emnlp-main.213} {{COMET}: {A} {Neural} {Framework} for {MT} {Evaluation}}.
\newblock In \emph{Proceedings of the 2020 {Conference} on {Empirical} {Methods} in {Natural} {Language} {Processing} ({EMNLP})}, pages 2685--2702, Online. Association for Computational Linguistics.

\bibitem[{Saleva and Lignos(2023)}]{saleva-lignos-2023-changes}
Jonne Saleva and Constantine Lignos. 2023.
\newblock \href {https://doi.org/10.18653/v1/2023.insights-1.7} {What changes when you randomly choose {BPE} merge operations? not much.}
\newblock In \emph{Proceedings of the Fourth Workshop on Insights from Negative Results in NLP}, pages 59--66, Dubrovnik, Croatia. Association for Computational Linguistics.

\bibitem[{Sennrich et~al.(2016)Sennrich, Haddow, and Birch}]{sennrich-etal-2016-neural}
Rico Sennrich, Barry Haddow, and Alexandra Birch. 2016.
\newblock \href {https://doi.org/10.18653/v1/P16-1162} {Neural machine translation of rare words with subword units}.
\newblock In \emph{Proceedings of the 54th Annual Meeting of the Association for Computational Linguistics (Volume 1: Long Papers)}, pages 1715--1725, Berlin, Germany. Association for Computational Linguistics.

\bibitem[{Uzan et~al.(2024)Uzan, Schmidt, Tanner, and Pinter}]{uzan_greed_2024}
Omri Uzan, Craig~W. Schmidt, Chris Tanner, and Yuval Pinter. 2024.
\newblock \href {https://doi.org/10.48550/arXiv.2403.01289} {Greed is {All} {You} {Need}: {An} {Evaluation} of {Tokenizer} {Inference} {Methods}}.
\newblock \emph{arXiv preprint}.
\newblock ArXiv:2403.01289 [cs].

\bibitem[{Yang et~al.(2024)Yang, Yang, Hui, Zheng, Yu, Zhou, Li, Li, Liu, Huang, Dong, Wei, Lin, Tang, Wang, Yang, Tu, Zhang, Ma, Yang, Xu, Zhou, Bai, He, Lin, Dang, Lu, Chen, Yang, Li, Xue, Ni, Zhang, Wang, Peng, Men, Gao, Lin, Wang, Bai, Tan, Zhu, Li, Liu, Ge, Deng, Zhou, Ren, Zhang, Wei, Ren, Liu, Fan, Yao, Zhang, Wan, Chu, Liu, Cui, Zhang, Guo, and Fan}]{yang2024qwen2technicalreport}
An~Yang, Baosong Yang, Binyuan Hui, Bo~Zheng, Bowen Yu, Chang Zhou, Chengpeng Li, Chengyuan Li, Dayiheng Liu, Fei Huang, Guanting Dong, Haoran Wei, Huan Lin, Jialong Tang, Jialin Wang, Jian Yang, Jianhong Tu, Jianwei Zhang, Jianxin Ma, Jianxin Yang, Jin Xu, Jingren Zhou, Jinze Bai, Jinzheng He, Junyang Lin, Kai Dang, Keming Lu, Keqin Chen, Kexin Yang, Mei Li, Mingfeng Xue, Na~Ni, Pei Zhang, Peng Wang, Ru~Peng, Rui Men, Ruize Gao, Runji Lin, Shijie Wang, Shuai Bai, Sinan Tan, Tianhang Zhu, Tianhao Li, Tianyu Liu, Wenbin Ge, Xiaodong Deng, Xiaohuan Zhou, Xingzhang Ren, Xinyu Zhang, Xipin Wei, Xuancheng Ren, Xuejing Liu, Yang Fan, Yang Yao, Yichang Zhang, Yu~Wan, Yunfei Chu, Yuqiong Liu, Zeyu Cui, Zhenru Zhang, Zhifang Guo, and Zhihao Fan. 2024.
\newblock \href {https://arxiv.org/abs/2407.10671} {Qwen2 technical report}.
\newblock \emph{Preprint}, arXiv:2407.10671.

\bibitem[{Zouhar et~al.(2023)Zouhar, Meister, Gastaldi, Du, Vieira, Sachan, and Cotterell}]{zouhar-etal-2023-formal}
Vil{\'e}m Zouhar, Clara Meister, Juan Gastaldi, Li~Du, Tim Vieira, Mrinmaya Sachan, and Ryan Cotterell. 2023.
\newblock \href {https://doi.org/10.18653/v1/2023.findings-acl.38} {A formal perspective on byte-pair encoding}.
\newblock In \emph{Findings of the Association for Computational Linguistics: ACL 2023}, pages 598--614, Toronto, Canada. Association for Computational Linguistics.

\end{thebibliography}
\appendix
\appendix
\section{Appendix}

% \input{data/figures/algorithms/bpe_train}
% Add this to your preamble:
\setlength{\tabcolsep}{4pt}
\begin{algorithm*}[ht]
\caption{Dynamic programming for maximal-compression BPE encoding. Given an input string $\mathrm{s}$ and a BPE vocabulary represented as a prefix trie rooted at \texttt{root}, this procedure finds the shortest sequence of token IDs whose concatenation exactly matches $\mathrm{s}$. We maintain a one-dimensional array \texttt{dp[0..n]} where \texttt{dp}[i] holds the best encoding (minimal number of tokens) for the prefix $\mathrm{s}[0..i-1]$. At each position $i$, we traverse the trie from the root to extend all valid tokens starting at $i$, updating \texttt{dp}[j+1] whenever we discover a shorter encoding ending at $j$. Time complexity $\mathcal{O}(n^2)$, space complexity $\mathcal{O}(n)$.}
\label{algo:maxcompbpe}
\centering
{\footnotesize
  \ra{1.3}  
    \begin{algorithmic}[1]
      \Procedure{MaxCompBPEEncode}{$\mathrm{s}, \mathrm{root}$}
        \State $\mathrm{n} \gets |\mathrm{s}|$
        \State $\mathrm{dp} \gets [\mathrm{None}]_{0..\mathrm{n}}$
        \State $\mathrm{dp}[0] \gets []$
        \For{$i \gets 0$ \textbf{to} $\mathrm{n}-1$}
          \If{$\mathrm{dp}[i]\neq \mathrm{None}$}
            \State $\mathrm{node}\gets \mathrm{root}$
            \For{$j \gets i$ \textbf{to} $\mathrm{n}-1$}
              \If{$\mathrm{s}[j]\notin \mathrm{node}.\mathrm{children}$}
                \State \textbf{break}
              \EndIf
              \State $\mathrm{node}\gets \mathrm{node}.\mathrm{children}[\mathrm{s}[j]]$
              \If{$\mathrm{node}.\mathrm{token\_id}$ is defined}
                \State $\mathrm{candidate}\gets \mathrm{dp}[i]\;\Vert\;\mathrm{node}.\mathrm{token\_id}$
                \If{$(\mathrm{dp}[j+1]=\mathrm{None})\;\lor\;|\mathrm{candidate}|<|\mathrm{dp}[j+1]|$}
                  \State $\mathrm{dp}[j+1]\gets \mathrm{candidate}$
                \EndIf
              \EndIf
            \EndFor
          \EndIf
        \EndFor
        \State \Return $\mathrm{dp}[\mathrm{n}]$
      \EndProcedure
    \end{algorithmic}
}
\end{algorithm*}

% \input{data/figures/algorithms/left_to_right}

% Add this to your preamble:
\setlength{\tabcolsep}{4pt}

\begin{figure*}[ht]
  \centering
  {\footnotesize
    \begin{adjustbox}{max width=\textwidth}
      \ra{1.3}
        \includegraphics[width=0.8\textwidth]{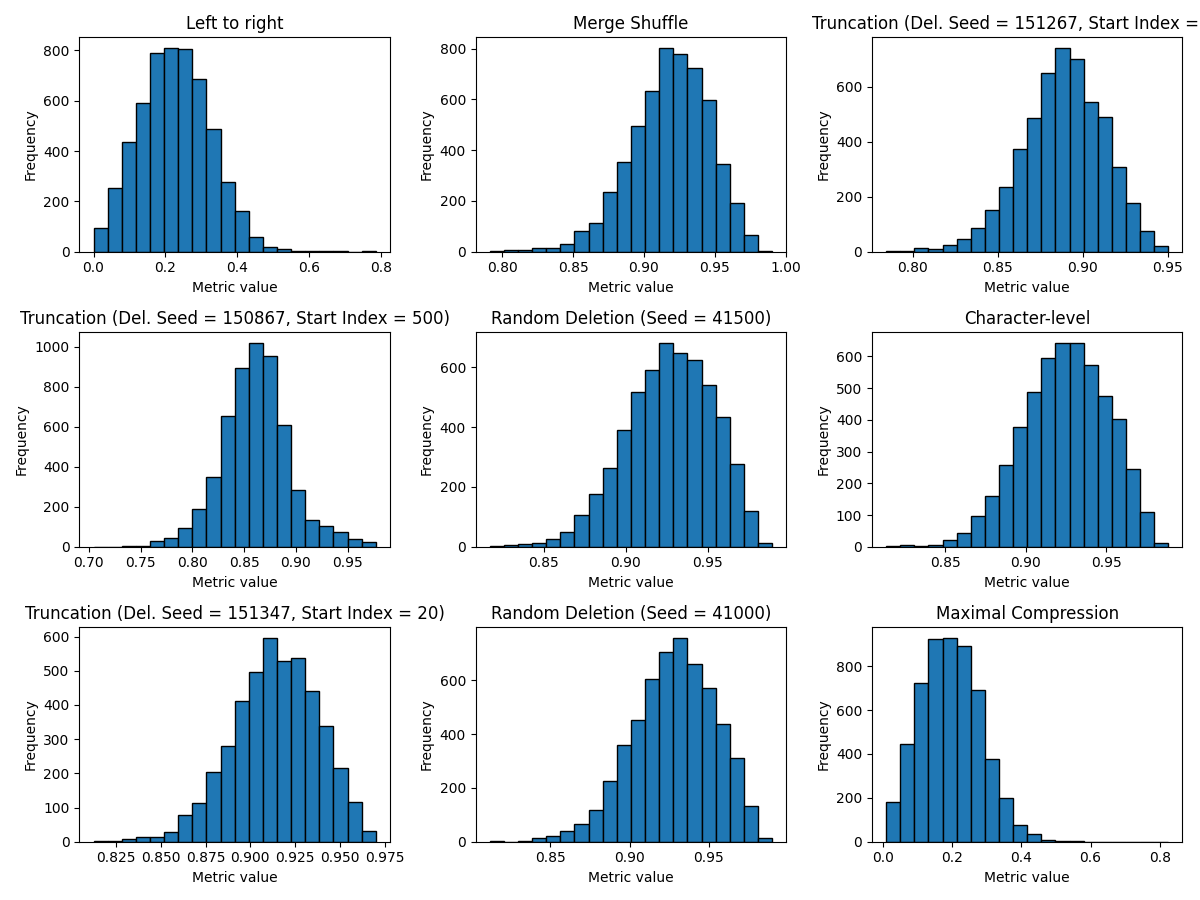}
    \end{adjustbox}
  }
  \caption{
    Jaccard distance between the tokenization of the Semantic Scholar prompts obtained from the standard tokenizer and custom tokenizers.  
  }
  \label{fig:jaccard}
\end{figure*}

% Add this to your preamble:
\setlength{\tabcolsep}{4pt}

\begin{figure*}[ht]
  \centering
  {\footnotesize
    \begin{adjustbox}{max width=\textwidth}
      \ra{1.3}
      \includegraphics[width=0.8\textwidth]{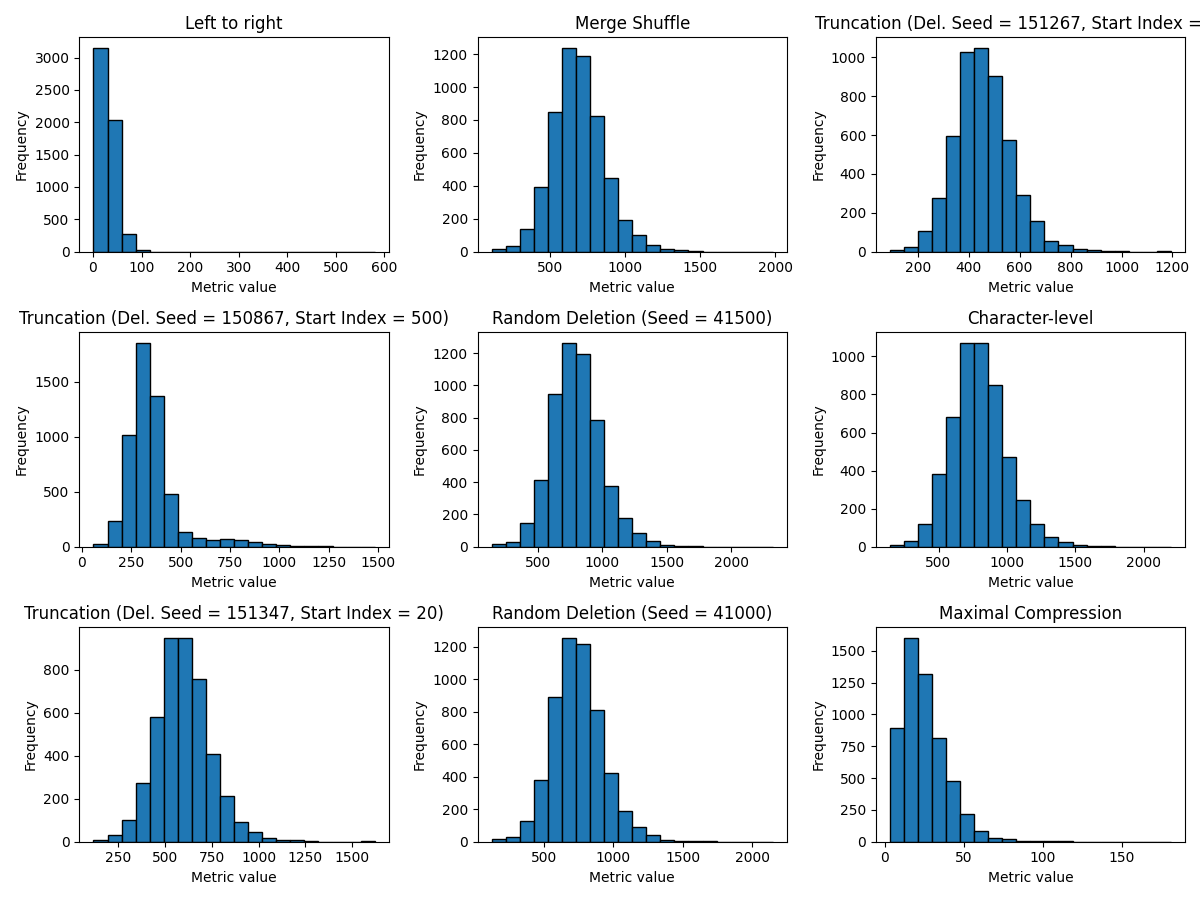}
    \end{adjustbox}
  }
  \caption{
    Levenshtein distance between the tokenization of the Semantic Scholar prompts obtained from the standard tokenizer and custom tokenizers.  
  }
  \label{fig:levenshtein}
\end{figure*}

% Add this to your preamble:
\setlength{\tabcolsep}{4pt}

\begin{figure*}[ht]
  \centering
  {\footnotesize
    \begin{adjustbox}{max width=\textwidth}
      \ra{1.3}
      \includegraphics[width=0.8\textwidth]{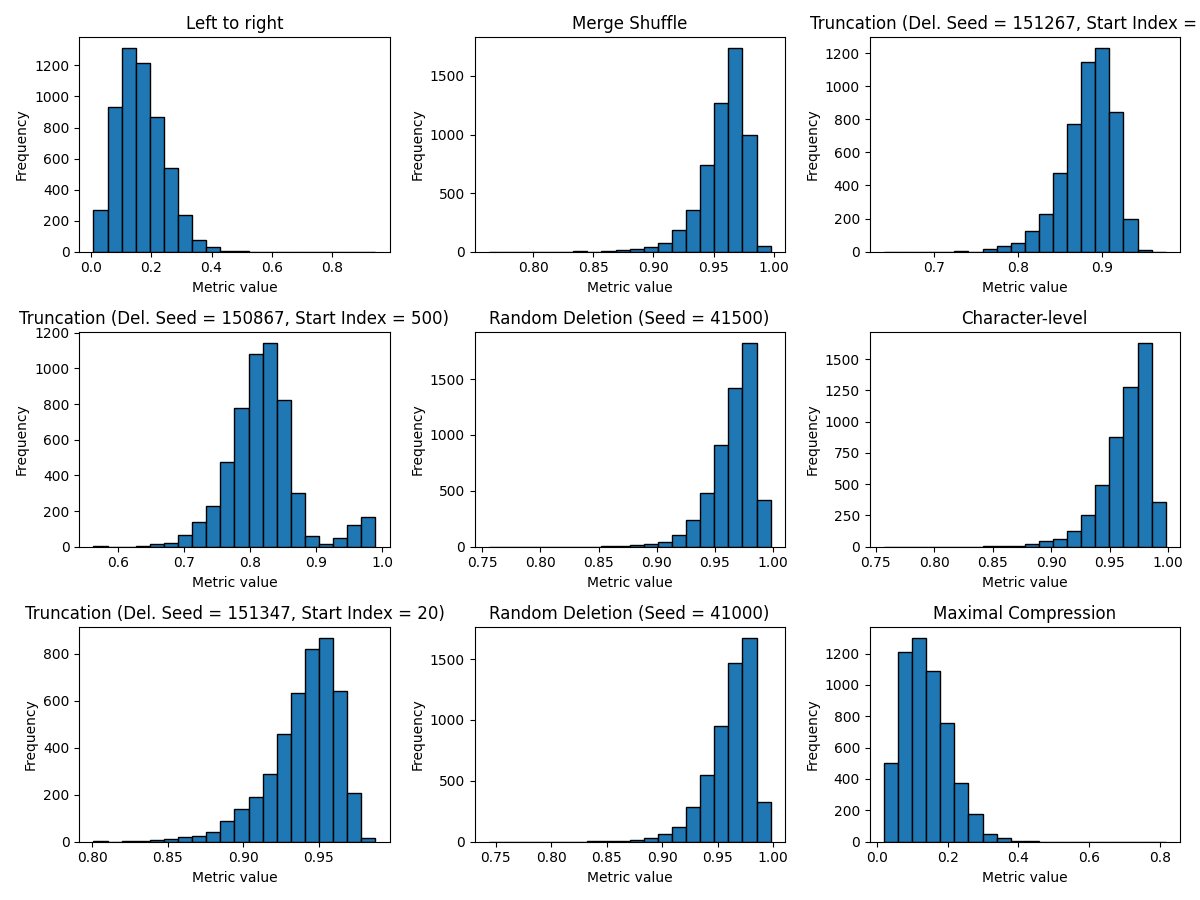}
    \end{adjustbox}
  }
  \caption{
    Token-level Levenshtein edit distance between the tokenization of S2ORC prompts obtained from the standard tokenizer and alternative tokenizers (open-ended generation subset). Lower is closer to standard.
  }
  \label{fig:edit-distance}
\end{figure*}

% Machine translation detailed metrics
% BLEU is computed on detokenized outputs
% Your table goes into the document body
\begin{table*}\centering
\ra{1.3}
\begin{tabular}{@{}lrrr@{}}\toprule
Tokenizer & COMET & xCOMET & CometKiwi \\ \midrule
Standard & 0.502 & 0.585 & 0.372 \\
Maximal Compression & 0.494 & 0.550 & 0.348 \\
Left to right & 0.495 & 0.567 & 0.394 \\
Merge Shuffle & 0.478 & 0.484 & 0.220 \\
Character-level & 0.479 & 0.509 & 0.253 \\
Random Deletion & 0.482 & 0.523 & 0.267 \\
\bottomrule
\end{tabular}
\caption{German-to-English machine translation results (COMET primary; xCOMET, CometKiwi in Appendix for completeness).}
\label{tab:de-to-en-comet}
\end{table*}

\begin{table*}
    \centering
    \ra{1.3}
    \footnotesize
    \begin{tabular}{@{}lrrrrr@{}}
    \toprule
    Tokenizer & COMET & xCOMET & CometKiwi & BLEU & METEOR \\
    \midrule
    Standard & 0.685 & 0.545 & 0.366 & 18.4 & 0.468 \\
    Maximal Compression & 0.633 & 0.454 & 0.296 & 13.0 & 0.412 \\
    Left to right & 0.632 & 0.457 & 0.298 & 12.8 & 0.413 \\
    Merge Shuffle & 0.632 & 0.263 & 0.140 & 3.49 & 0.210 \\
    Character-level & 0.519 & 0.279 & 0.171 & 5.23 & 0.240 \\
    Random deletion & 0.531 & 0.288 & 0.182 & 5.32 & 0.251 \\
    \bottomrule
    \end{tabular}
    \caption{Czech-to-English machine translation results. BLEU is computed on detokenized outputs.}
    \label{tab:tokenizer-comparison}
\end{table*}

% UPDATE 0117-1700: Encoding details moved to background section. 
% \section{Appendix: Encoding details for popular BPE implementations.}
% \input{sections/appendix/encoding_details.tex}

%%%%%%%%%%%
% ACL latex template fails to compile if document is less than 1 page long. 
% Use loren ipsum to fill in the document. 
% \newpage
% \lipsum[1-5]

\end{document}